# Risk-aware Trajectory Prediction by Incorporating Spatio-temporal Traffic Interaction Analysis

Divya Thuremella, Lewis Ince and Lars Kunze

*Abstract*— To operate in open-ended environments where humans interact in complex, diverse ways, autonomous robots must learn to predict their behaviour, especially when that behavior is potentially dangerous to other agents or to the robot. However, reducing the risk of accidents requires prior knowledge of where potential collisions may occur and how. Therefore, we propose to gain this information by analyzing locations and speeds that commonly correspond to high-risk interactions within the dataset, and use it within training to generate better predictions in high risk situations. Through these location-based and speed-based re-weighting techniques, we achieve improved overall performance, as measured by most-likely FDE and KDE, as well as improved performance on high-speed vehicles, and vehicles within high-risk locations.

## I. INTRODUCTION

Vulnerable road users (VRUs), i.e., pedestrians, cyclists, motor cyclists, and horse riders [1], account for more than half of all road traffic accident related deaths worldwide [2] and are the hardest to predict since pedestrians are not restricted to the use of lanes as vehicles are [3]. Furthermore, every 1% increase in mean speed produces a 4% increase in fatal crash risk and 3% increase in serious crash risk [2], indicating that high speed vehicles pose a higher risk to VRUs and to other cars around them. Hence, high-risk pedestrian-vehicle interactions and high speed vehicles are two major sources of accidental collision.

Predicting collisions and the behaviors used to prevent imminent collision can help autonomous robots to anticipate unexpected trajectories of near-collision situations and plan a safe path accordingly [4]. Risk/collision prediction can also help road planners improve road safety and traffic congestion [5]. Therefore, we aim to improve trajectory prediction in high-risk situations, specifically, those involving the previously discussed sources of accidental collision: close pedestrian-vehicle interactions and high speed vehicles.

Furthermore, we recognize that many trajectory prediction datasets include a large number of stationary vehicles which are easier to predict and therefore may bias models and evaluation metrics to show superficially better performance. In this work, we report separate metrics for performance on stationary and non-stationary vehicles, so that better understanding of the model's performance can be gained.

Our contributions include: 1) identifying areas of the map with high-risk pedestrian-vehicle interactions, 2) improving model performance on those high-risk areas, 3) additionally improving performance for dangerous high-speed vehicles, and 4) extensively analyzing both the baseline and our models on stationary and non-stationary vehicles, high-speed vehicles, and road users within high-risk locations.

In this work, we study the risk that observed road users pose towards each other, and aim to understand how interactions between them change due to the presence of risk conditions. We do not define risk as a measured uncertainty of prediction confidence or as the danger autonomous robots may pose to humans. The code [1] for our work is available.

## II. RELATED WORK.

### A. Spatially Aware Trajectory Prediction

Many trajectory prediction algorithms use map information to add spatial context to the prediction model (e.g. [6]–[9]). However, some models add extra context (calculated heuristically or through model learning) to locations on the map. For example, [10] combines semantic map labels and positions of social agents into an occupancy map with a sliding scale: agents, cones, and curbs are high occupancy, merge lanes are medium occupancy (since they're likely to soon be occupied), and fast lanes are low occupancy. Similarly, [11] creates an occupancy map from lane lines and road boundaries. Some methods, on the other hand, dedicate part of their model towards learning extra spatial information: [12] produce a heatmap of potential goal locations from the map and agent history to better inform the prediction model while [13] learn latent map features using an autoencoder to provide latent spatial information.

However, one underexplored area of spatially aware trajectory prediction is the study of spatio-temporal patterns across the dataset and creation of location-specific priors based on these patterns, to incorporate into trajectory prediction. One implementation of this technique is the building of MoDs (maps of dynamics), which gather spatial velocity information from the training data and encode learned stochastic motion patterns at each location [14]. This is used to bias the prediction model [14]. Another method which improves prediction using spatio-temporal pattern priors is [15]. [15] extracts the spatial risk pattern at each timestep in the traffic scene by calculating the TTC (time-to-collision) between every pair of vehicles, records the locations with low TTC, and creates a spatial risk heatmap by using the average inverse TTC value at every location. This approach is the most similar to our work, with two primary differences. Firstly, [15]

The authors are with the Oxford Robotics Institute, Department of Engineering, Oxford University, Oxford, UK divya@robots.ox.ac.uk, lars@robots.ox.ac.uk.
This work was supported by the EPSRC project RAILS (grant reference: EP/W011344/1), and the Oxford Robotics Institute research project RobotCycle.

[1] https://github.com/cognitive-robots/risk-aware-trajectory-prediction

only consider vehicle-vehicle interactions, and evaluate on the highway datasets NGSIM and HighD, whereas we focus on pedestrian-vehicle interactions and evaluate on the diverse city environments in NuScenes. Secondly, [15] use the spatial heatmap as an input to their model, while we use its values to directly re-weight the loss. While TTC is a more well-studied measure of risk than the simple minimum distance method used in our work, it cannot be applied to pedestrian-vehicle interactions because pedestrians can have large lateral motion [16]. Furthermore, [15] only show overall evaluation metrics while we also show evaluations centered on high-risk examples.

*B. Risk-Aware Trajectory Prediction*

While there are many trajectory prediction methods which use a collision risk formulation to prevent the prediction of trajectories which collide with each other (e.g. [6], [17], [18]), we aim to increase understanding and performance in situations where agents do collide (or almost collide) so that high-risk potential collision interactions can be better predicted. Therefore, we discuss here the methods which use non-spatial calculations of risk (as spatial methods are covered above) to improve performance on high-risk trajectories.

[16] propose a risk metric applicable to both vehicles and VRUs which exhibit large lateral motion and combine this with a trajectory prediction model to create a predictive risk analysis model which can accurately forecast risky scenarios 3 seconds before they happen. Although we use a simple distance-based risk estimation in this work, the future direction of our research aims to use the risk metric defined by [16] within the risk-aware prediction method of this work.

[19] and [20] use CVaR (probability of predictions below a certain error) as the formulation of risk, which aims to reduce prediction error of the worst performing long-tail of the dataset. [20] is similar to our method in that it performs re-weighting to increase importance of high-risk examples.

III. METHODOLOGY

**Dataset.** To train and evaluate our model, we use the NuScenes dataset [21], which consists of 1000 scenes with 5.5 hours of footage labeled at 2Hz. It has 17,081 labeled tracks taken from a moving vehicle in 4 neighborhoods within Boston and Singapore (boston-seaport, singapore-onenorth, singapore-queenstown, singapore-hollandvillage), and includes HD semantic maps with 11 annotated layers, including pedestrian crossings, walkways, stop lines, traffic lights, road dividers, lane dividers, and driveable areas [21].

**Baseline**. We use the same training methodology and learning parameters as our baseline, Trajectron++ [8], with 12 epochs, data augmentation, dynamics integration, and semantic maps. The Trajectron++ [8] architecture encodes the semantic map, history, and social interactions via a CNN and LSTMs respectively, and puts the concatenated input feature encodings into a CVAE to learn a latent space embedding, which is then used to predict a multi-modal distribution of likely future positions iteratively using a gaussian mixture model on top of a GRU.

*A. Location-Risk Model*

Data shows that in busy intersections and shopping areas, risk of pedestrian-vehicle collision is much greater [22]. While such areas can be labeled or inferred from a semantic map like NuScenes [21], maps lack the spatio-temporal information that residents would know: knowledge of which intersections and shopping areas are busier and when. Fortunately, this information can be gathered from patterns in training data [15].

To gather these patterns, we first determine the distance between every non-stationary vehicle and VRU within each scene (since interactions between VRUs and non-stationary vehicles pose the highest risk), and discretize the map of each neighborhood in NuScenes into a 100x100 grid. Then, we assign a risk score to each of the 10k bins of the grid in the following way: for each timestep, the value of the location bin containing the closest (distance-wise) vehicle-VRU interaction in the map is incremented. This creates a 100x100 2D histogram that shows the per-location frequency of the number of timesteps at which the riskiest vehicle-VRU interaction of the timestep had that location. Finally, this histogram is normalized, shifted, and re-scaled such that the minimum value is 1 and the maximum value is 10 (in order to give a re-weighting factor of 1 to low-risk locations and 10 to high-risk locations). The resulting heatmap is shown in Fig. 1b. As can be seen, the highest heatmap values correspond to busy intersections and streets with a high concentration of offices and restaurants (as informed by Google Maps). The heatmap values are then used as the re-weighting factor by which examples in each location bin are weighted.

One caveat, however, is that since pedestrians have right-of-way in intersections with crosswalks (commonly found in the NuScenes dataset), we expect pedestrian behavior in high-risk areas to be similar to their behavior in other areas (i.e. maintaining a steady speed across the road, as well as when walking down sidewalks). Vehicles, which have to watch for pedestrians and stop when one is crossing, significantly change their behavior in busy intersections. Therefore, we hypothesize that pedestrian performance in high-risk locations like busy intersections will remain relatively unchanged by this model, while vehicle prediction performance will improve due to the added risk information.

*B. Non-Stationary Model*

As expected, the Location-Risk Model improved vehicle performance on average and in high-risk locations. However, prediction of high speed vehicles (i.e. vehicles whose maximum speed reaches $> 14m/s$) regressed (see Table I). Since high speed vehicles are a major source of risk, we aim to improve prediction of high speed vehicles while maintaining the boosted performance of vehicles in high-risk areas.

Therefore, we analyzed the distribution of vehicle speeds in the training data (see Fig. 2c) and found that 27% of vehicles had a maximum speed of 0. One possible reason for this is that within the dataset preprocessing of Trajectron++ [8], vehicles with a total path length (i.e. distance traveled throughout the total time observed in the scene) of less than 1m were considered to be stationary and their velocities were smoothed to 0m/s. Firstly, this poses a data diversity issue: if 27% of the examples are almost identical in trajectory, the model may be over-fitting to these examples [23]. This is similar to the class imbalance problem in classification, but since stationary vehicles are easier to predict, this imbalance may also bias the evaluation to be easier. Secondly, this can be considered data leakage: since the velocities of vehicles which start moving after some time are not smoothed, the model may be 'cheating' by recognizing smoothed trajectories as stationary. Therefore, we calculate stationary and non-stationary vehicle performance separately (see Table I) and compare the baseline model's performance on these two classes of vehicles. Furthermore, to observe what happens to performance when stationary vehicles are removed, we train a model to apply a re-weighting factor of 0 to stationary vehicles and 1 to non-stationary vehicles, such that stationary vehicles do not contribute to the loss.

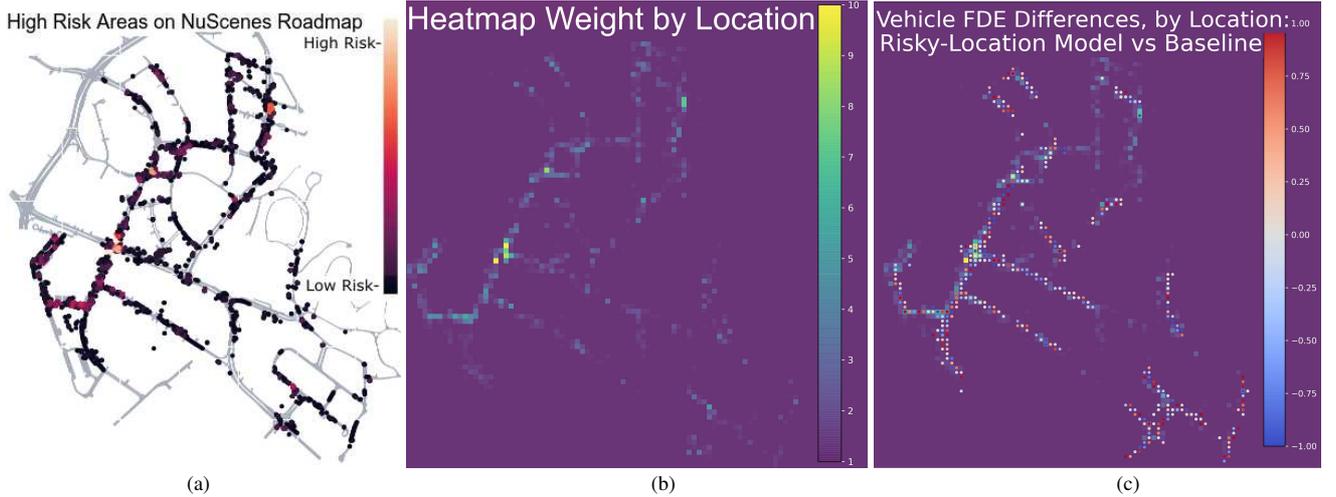

Fig. 1. (a) areas on the 'singapore-onenorth' NuScenes Map that were surfaced as high-risk (like big intersections) based on our risk criteria. This information was used to create the heatmap (b), which determines (on a scale of 1 to 10) the weight that examples in each corresponding location get. After training/testing the Location-Risk and Location-Risk+Non-Stationary models, we plotted the average improvement/regression at each heatmap location as compared to the baseline (c). Red dots indicates regression and blue dots indicate improvements over the baseline, as measured by the FDE. The colorbar shows the exact number (in meters) of FDE difference from baseline. Key Takeaway: Locations with highest risk (i.e. yellow heatmap squares and orange dots on roadmap) show location-wise improvements over the baseline (since dots overlayed on those locations are largely blue).

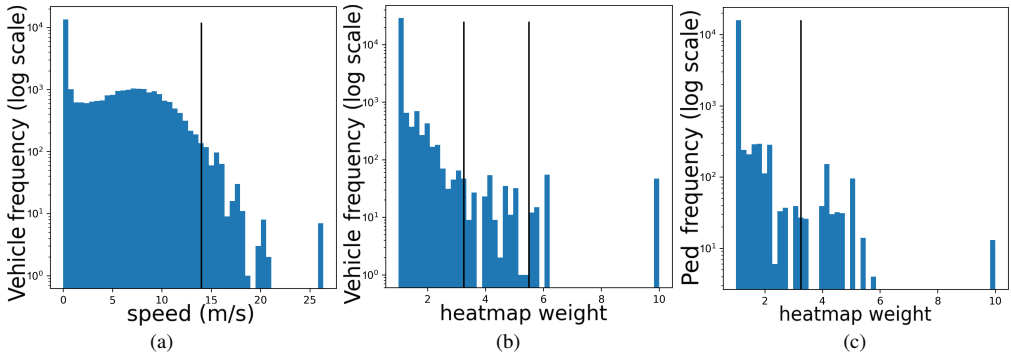

Fig. 2. (a) distribution of vehicle speeds in test set to show how many examples fall within the stationary (0m/s), non-stationary, and high ($> 14m/s$) speed ranges. (b) distribution of location-risk weights given in test set to show how many examples fall within the low, medium, and high-risk categories via the black lines. Higher heatmap weights indicate higher risk. (c) test set distribution of location-risk weights given to pedestrians showing the split into low and high-risk categories via the black line.

### C. Location-Risk+Non-Stationary Model

Since removing stationary vehicles' effect on the loss in the Non-Stationary model showed improved performance on high speed vehicles (Fig. 3a), we combine the two models by both weighting stationary vehicles by 0, and weighting high-risk locations according to the heatmap in Fig. 1b. We hypothesize that this model will improve prediction for high speed vehicles as well as vehicles in high-risk areas.

## IV. EVALUATION METRICS

Though the model was only trained to predict 3s into the future, we evaluate on predictions that are 1,2,3, and 4s into the future, as done in [8] to demonstrate ability to generalize.

We follow most methods using NuScenes (e.g. [8], [24], [25]) and use the final distance error (FDE) of the most likely predicted trajectory as our main evaluation metric, but unlike

TABLE I

COMPARE PERFORMANCE ON THE FOLLOWING SPEED CATEGORIES: STATIONARY VEHICLES (STAT), NON-STATIONARY VEHICLES (NO-STAT), AND HIGH SPEED VEHICLES (ABOVE SPEED LIMIT 14M/S). ALL INDICATES AVERAGE PERFORMANCE OF ENTIRE TEST SET. METRIC USED IS MOST-LIKELY FDE ON NUSCENES. KEY TAKEAWAY: NON-STATIONARY MODEL IMPROVES HIGH-SPEED VEHICLE PREDICTION AND LOCATION-RISK+NON-STATIONARY MODEL SHOWS HIGHEST PERFORMANCE ON NON-STATIONARY VEHICLES.

| Model | Vehicle FDE (m) @1s | | | | Vehicle FDE (m) @2s | | | | Vehicle FDE (m) @3s | | | | Vehicle FDE (m) @4s | | | |
|---|---|---|---|---|---|---|---|---|---|---|---|---|---|---|---|---|
| | all | stat | no-stat | $>14m/s$ | all | stat | no-stat | $>14m/s$ | all | stat | no-stat | $>14m/s$ | all | stat | no-stat | $>14m/s$ |
| Baseline | 0.07 | 0.02 | 0.10 | 0.29 | 0.45 | **0.04** | 0.71 | 1.19 | 1.14 | 0.07 | 1.83 | 2.57 | 2.21 | 0.14 | 3.54 | 4.51 |
| Location-Risk | **0.06** | **0.01** | **0.09** | 0.29 | **0.43** | **0.04** | 0.68 | 1.21 | **1.08** | **0.06** | 1.74 | 2.58 | 2.08 | 0.13 | 3.33 | 4.47 |
| Non-Stationary | **0.06** | 0.02 | **0.09** | 0.29 | 0.44 | 0.05 | 0.68 | 1.19 | 1.10 | 0.08 | 1.76 | **2.48** | 2.08 | 0.10 | 3.35 | **4.23** |
| Location-Risk+ Non-Stationary | **0.06** | 0.02 | **0.09** | 0.29 | **0.43** | 0.05 | **0.68** | **1.17** | 1.09 | 0.08 | **1.73** | 2.50 | **2.03** | **0.09** | **3.28** | 4.30 |

TABLE II

COMPARE PERFORMANCE ACROSS DIFFERENT TYPES OF LOCATIONS (CATEGORIES BASED ON THEIR RISK FACTOR). METRIC USED IS MOST-LIKELY FDE ON NUSCENES. L, M, AND H INDICATE PERFORMANCE ON LOW-RISK, MEDIUM-RISK, AND HIGH-RISK LOCATIONS. KEY TAKEAWAY: THE LOCATION-RISK MODEL MAKES THE HIGHEST IMPROVEMENTS IN HIGH-RISK LOCATIONS, WHILE STILL IMPROVING METRICS ACROSS THE BOARD.

| Model | Vehicle FDE @1s | | | | Vehicle FDE @2s | | | | Vehicle FDE @3s | | | | Vehicle FDE @4s | | | |
|---|---|---|---|---|---|---|---|---|---|---|---|---|---|---|---|---|
| | all | L | M | H | all | L | M | H | all | L | M | H | all | L | M | H |
| Baseline | 0.07 | 0.07 | 0.10 | 0.07 | 0.45 | 0.45 | 0.65 | 0.46 | 1.14 | 1.13 | 1.65 | 1.32 | 2.21 | 2.19 | 3.14 | 2.79 |
| Location-Risk | **0.06** | **0.06** | **0.09** | **0.06** | **0.43** | **0.43** | 0.63 | **0.40** | **1.08** | **1.08** | 1.58 | **1.12** | 2.08 | 2.06 | 3.01 | **2.30** |
| Non-Stationary | **0.06** | **0.06** | 0.10 | **0.06** | 0.44 | **0.43** | 0.63 | 0.41 | 1.10 | 1.09 | 1.61 | 1.15 | 2.08 | 2.07 | 3.06 | 2.37 |
| Location-Risk+ Non-Stationary | **0.06** | **0.06** | **0.09** | **0.06** | **0.43** | **0.43** | **0.62** | 0.41 | 1.09 | **1.08** | **1.53** | 1.19 | **2.03** | **2.02** | **2.88** | 2.56 |

TABLE III

COMPARE KDE ACROSS DIFFERENT TYPES OF LOCATIONS, WHERE L, M, AND H INDICATE PERFORMANCE ON LOW-RISK, MEDIUM-RISK, AND HIGH-RISK TYPES OF LOCATIONS. KEY TAKEAWAY: ALL 3 MODELS MAKE ACROSS-THE-BOARD IMPROVEMENTS OVER THE BASELINE ON KDE, WHICH TAKES ENTIRE PREDICTED DISTRIBUTION INTO ACCOUNT, NOT JUST MOST-LIKELY FDE.

| Model | Vehicle KDE @1s | | | | Vehicle KDE @2s | | | | Vehicle KDE @3s | | | | Vehicle KDE @4s | | | |
|---|---|---|---|---|---|---|---|---|---|---|---|---|---|---|---|---|
| | all | L | M | H | all | L | M | H | all | L | M | H | all | L | M | H |
| Baseline | -4.18 | -4.19 | -2.84 | -3.74 | -2.74 | -2.76 | -1.18 | -2.05 | -1.61 | -1.64 | 0.00 | -0.78 | -0.71 | -0.73 | 0.95 | 0.35 |
| Location-Risk | -4.21 | -4.23 | **-2.97** | **-3.95** | -2.80 | -2.82 | -1.27 | **-2.30** | **-1.68** | **-1.70** | -0.09 | **-0.92** | **-0.77** | **-0.79** | 0.89 | **0.18** |
| Non-Stationary | **-4.26** | **-4.28** | -2.95 | -3.90 | **-2.82** | **-2.84** | **-1.31** | -2.18 | **-1.68** | **-1.70** | **-0.10** | -0.86 | **-0.77** | **-0.79** | 0.87 | 0.23 |
| Location-Risk+ Non-Stationary | -4.21 | -4.22 | -2.93 | -3.91 | -2.76 | -2.78 | -1.30 | -2.16 | -1.63 | -1.65 | -0.08 | -0.82 | -0.72 | -0.74 | **0.86** | 0.23 |

TABLE IV

COMPARE PEDESTRIAN PERFORMANCE ACROSS TYPES OF LOCATIONS (L, M, AND H INDICATE PERFORMANCE ON LOW-RISK, MEDIUM-RISK, AND HIGH-RISK TYPES OF LOCATIONS). KEY TAKEAWAY: SMALL IMPROVEMENTS ARE MADE ALMOST ACROSS THE BOARD BY THE LOCATION-RISK MODEL, AND THOUGH THE NON-STATIONARY MODEL REGRESSES PERFORMANCE ACROSS THE BOARD, THESE REGRESSIONS ARE REVERSED BY ADDITION OF THE LOCATION-RISK LOSS IN THE LOCATION-RISK+NON-STATIONARY MODEL.

| Model | Pedestrian FDE @1s | | | Pedestrian FDE @2s | | | Pedestrian FDE @3s | | | Pedestrian FDE @4s | | |
|---|---|---|---|---|---|---|---|---|---|---|---|---|
| | all | L | H | all | L | H | all | L | H | all | L | H |
| Baseline | **0.014** | **0.014** | **0.015** | 0.166 | 0.165 | 0.197 | 0.370 | 0.367 | 0.472 | 0.618 | 0.612 | 0.856 |
| Location-Risk | 0.019 | 0.019 | 0.019 | **0.164** | **0.163** | **0.193** | **0.362** | **0.360** | **0.463** | **0.602** | **0.596** | **0.829** |
| Non-Stationary | 0.018 | 0.018 | 0.019 | 0.168 | 0.168 | 0.195 | 0.375 | 0.373 | 0.473 | 0.629 | 0.623 | 0.852 |
| Location-Risk+ Non-Stationary | 0.019 | 0.019 | 0.019 | **0.164** | **0.163** | **0.193** | **0.362** | **0.360** | **0.463** | **0.602** | **0.596** | **0.829** |

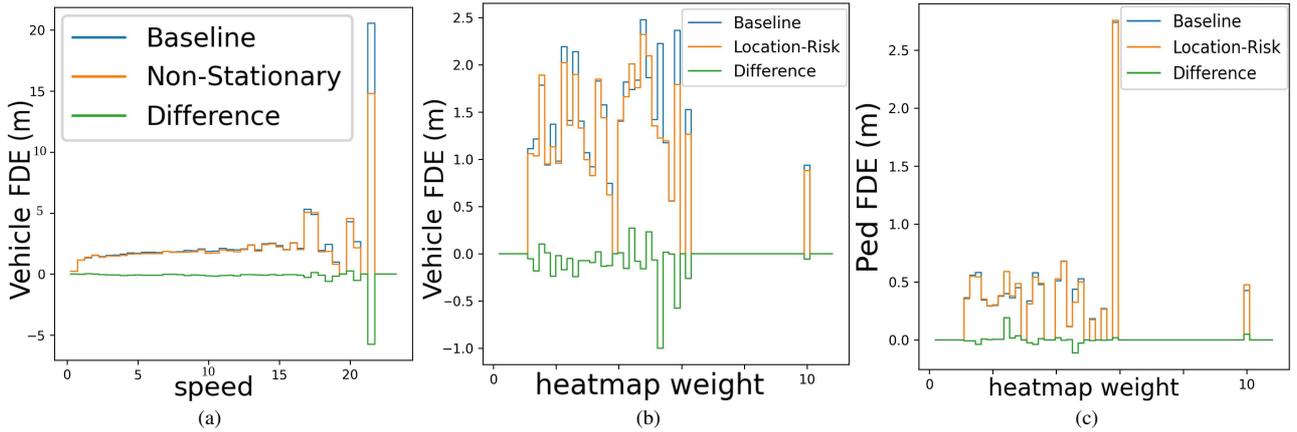

Fig. 3. (a) How improvements in performance vary for vehicles at different speeds: per-speed differences in most-likely FDE @3s between the Non-Stationary model and the baseline [8] to show how high speed vehicles are the most improved (i.e. most negative 'Difference' value) while low speed vehicles perform the same as baseline. (b) How improvements in performance vary across location-type: higher heatmap weight (higher risk) locations have generally bigger performance improvements (i.e. more negative difference values) than lower weight (lower risk) locations for vehicles, but no strong trend is seen for pedestrians, as shown by the per-location-type differences in most-likely FDE @3s between the Location-Risk model and baseline [8] in (c).

other methods, we focus on one type of metric so that we can use our limited space to show high-risk performance splits (i.e. performance on high-risk locations or high speed vehicles). To bolster our results, however, we also evaluate our models on the KDE-NLL metric used in [8] to show that performance of the entire predicted trajectory distribution is improved, and not just that of the most likely final prediction. KDE NLL is the mean negative log-likelihood of the ground truth trajectory using the probability density function of a distribution found by fitting a kernel density estimate on trajectory samples [26]. Therefore, it takes into account the full trajectories of the multimodal predicted distribution.

**Location-Specific Metrics**. To show improvements in high-risk locations, we define the following location-type metrics: we split the range of heatmap weights (the risk measure ranging from 1 to 10) into quarters, as shown in Fig. 2b, and aim to assess performance on the examples whose heatmap weights fall into each quarter. Therefore, the first quarter would contain low-risk examples whose location weights are $< 1.25$ and the last quarter would contain high-risk examples with location weights of $> 7.75$. However, there are very few vehicles whose weights fall into the last quarter, and very few pedestrians with weights in the last two quarters. Evaluating on sets of examples with such low numbers could cause performance instability. Therefore, we define the following location-based risk metrics: vehicles whose heatmap weights fall into the first quarter are considered 'low-risk', those whose weights fall into the second quarter are considered 'medium-risk', and those whose weights fall into the third and fourth quarters combined are considered 'high risk'. Similarly, pedestrians whose heatmap weights fall into the first quarter are considered 'low-risk', while those whose weights fall into the second, third, and fourth quarters combined, are 'high-risk,' (see Fig. 2c).

**Velocity-Specific Metrics**. Since the data is taken from both Boston, where the speed limit is 11 m/s [27], and Singapore, where the speed limit is 14 m/s [28], we use the maximum of these two speed limits and consider high speed vehicles to be those whose maximum speed throughout their history reaches above 14 m/s. Furthermore, since the baseline model may be biased towards stationary vehicles due to data leakage and low data diversity, we also report metrics on stationary and non-stationary vehicles separately. For this metric, we define stationary vehicles to be those whose path length from the first history to last future timestep is $< 1m$, and non-stationary vehicles to be all others.

**Color Plots**. In addition, we use color plots, as shown in Fig. 1, to demonstrate how the proposed models change the performance of examples in each location bin. For each bin, we average the performances of all vehicles in the test set whose current position lies within that bin, and subtract the per-location performance of baseline model from that of the Location-Risk model. As shown in the colormap, the colors range from FDE differences of -1 to 1 and show improvements in blue and regressions in red. FDE differences which are less or greater than -1 and 1 use the same shade of blue or red as -1 and 1.

**Stratified Histogram**. We also demonstrate how performance varies across the range of heatmap weights (Fig. 3) to show how a particular model impacts examples in high-risk locations, and vehicles of high speeds.

## V. RESULTS AND DISCUSSION

**Stationary vs. Non-Stationary Performance**. Since stationary vehicles are much easier to predict than non-stationary vehicles, we report the performance of the baseline model on the two classes separately. As shown in Table I, stationary vehicles have much better performance than non-stationary vehicles, especially at larger timeframes like @3s and @4s, where the performance is more than 25 times better. This shows how much the performance can be skewed when a large portion of the data is stationary. Furthermore, removal of stationary vehicles in the Non-Stationary model improved overall performance, showing that removing this skew can improve the model's general prediction ability.

### A. Location-Risk Model

As hypothesized, the Location-Risk model improves vehicle predictions, especially in high-risk locations, but shows only a little improvement in pedestrian predictions, even in high-risk locations (Fig. 3). Both the vehicle FDE metrics in Table II and vehicle KDE metrics in Table III show that the Location-Risk model has better performance across the board compared to the baseline. These tables show that the Location-Risk model doesn't simply improve vehicle prediction in only high-risk locations at the expense of other examples; it provides additional information to the model that improves average prediction overall.

The Location-Risk model shows better FDE performance on both stationary and non-stationary vehicles (see Table I), as well as improvement in the percentage of Road Boundary Violations, an indicator of improved understanding of the map [29]. One possible explanation for these improvements is that applying higher weights on examples within risky locations may be helping the model to better understand the semantic map. By assigning more importance to locations with high-risk interactions, the model may be required to use more information from the map to learn to differentiate high-risk situations and better predict within them. This improved understanding of the map may give the Location-Risk model a better ability to differentiate between stationary and non-stationary vehicles, thereby improving the prediction of both.

Pedestrian performance, however, shows only slight improvement in FDE (Table IV), and high-risk locations don't consistently show more improvement than low-risk locations (see Fig. 3). As hypothesized, this may be due to the fact that pedestrians with right-of-way may not change their behavior for vehicles as vehicles must for pedestrians.

Finally, in order to better understand the Location-Risk model and ensure that it is risk-aware for both high-risk locations, and high-risk agents, we also evaluate the model on high speed vehicles, as shown in Table I, and find that the Location-Risk model shows worse performance on high speed vehicles for some metrics (at 2s and 3s). Therefore, we add an additional re-weighting component designed to improve performance of such examples.

### B. Non-Stationary Model

Before combining the speed-based re-weighting component with the Location-Risk model, we first implement it on its own and compare its performance to all other models. When compared to the baseline model in Table I, the Non-Stationary model shows mostly regressions on stationary vehicle performance and across-the-board improvements on non-stationary vehicle performance (as to be expected), as well as small improvements to the high-speed vehicle performance. The graphs in Fig. 3a show a more nuanced story: although the average performance for speeds of $> 14m/s$ doesn't show much improvement, this model significantly improves the performance of vehicles with a speed of $> 20m/s$. One caveat, however, is that there is very little data that falls into this extremely high speed category, which can cause performance instability between model iterations.

One surprising result shown by Table III is that the Non-Stationary model is the best-performing out of the three proposed methods on overall KDE metrics. Since KDE is a metric that measures multimodality as well as accuracy, this shows that the Non-Stationary model may be improving multimodality in predictions. These results indicate that the removal of stationary vehicles may induce the model to learn more about the diversity of possible trajectories.

### C. Location-Risk+Non-Stationary Model

Combining the re-weighting schemes of the Location-Risk and Non-Stationary models shows slightly worse vehicle FDE and KDE performance on high-risk locations than the Location-Risk model alone, but the results are still an improvement over the baseline across the board. Furthermore, this model shows better performance on non-stationary vehicles and high-speed vehicles than the Location-Risk model, showing that our goal of improving the performance of high-speed vehicles while maintaining improved performance on high-risk locations was achieved.

Moreover, comparing the medium-risk improvements of the Location-Risk+Non-Stationary model to that of the Location-Risk model in Table II shows that the Location-Risk+Non-Stationary model yields even stronger improvements in medium-risk areas.

From these evaluations, it can be seen that the Non-Stationary and Location-Risk methods complement each other to bring higher vehicle performance both within high-risk locations and for high-risk (i.e. high speed) agents.

Pedestrian performance, on the other hand, is unchanged by the addition of the Non-Stationary model's re-weighting, as shown in Table IV. This is to be expected, as the re-weighting only applied to non-stationary vehicles.

## VI. CONCLUSION

Although there are many studies which calculate risk based on current kinematic information of surrounding agents [16] or current semantic map information [10], there is very little work done on examining the training data to gain prior knowledge of where, location-wise, collisions are likely to occur [15]. Therefore, we employ this strategy to gain advance knowledge of the likely locations of close pedestrian-vehicle interactions, and use it to improve the model's understanding of pedestrian and vehicle behavior. Using these spatio-temporal priors generates better prediction within those areas, and helps the model to better utilize the semantic map it has been given. Furthermore, we show that a high number of stationary examples in the dataset biases the model, and that removing stationary examples' effect on the loss improves overall performance, and especially the performance of high-speed vehicles. The results of our Location-Risk+Non-Stationary model show improved performance on both high-risk locations and high-speed vehicles.